\documentclass{article}
\usepackage{spconf,amsmath,graphicx, amssymb, amsfonts, amsthm, bm, algorithm2e, comment}

\usepackage{cite}
\usepackage{amsmath, amssymb}
\usepackage{algorithmic}
\usepackage{graphicx}
\usepackage{subfigure}

\usepackage{textcomp}
\usepackage{bm}

\usepackage[caption=false,font=normalsize,labelfont=sf,textfont=sf]{subfig}

\def\BibTeX{{\rmB\kern-.05em{\sci\kern-.025emb}\kern-.08emT\kern-.1667em\lower.7ex\hbox{E}\kern-.125emX}}

\begin{document}
\title{OPTIMAL COST CONSTRAINED ADVERSARIAL ATTACKS FOR MULTIPLE AGENT SYSTEMS}
\name{$^1$Ziqing Lu, $^2$Guanlin Liu, $^2$Lifeng Lai, $^1$Weiyu Xu}
\address{$^1$University of Iowa, $^2$University of California, Davis}
\maketitle
\begin{abstract}
Finding optimal adversarial attack strategies is an important topic in reinforcement learning and the Markov decision process.  Previous studies usually assume one all-knowing coordinator (attacker) for whom attacking different recipient (victim) agents incurs uniform costs. However, in reality, instead of using one limitless central attacker, the attacks often need to be performed by distributed attack agents. We formulate the problem of performing optimal adversarial agent-to-agent attacks using distributed attack agents, in which we impose distinct cost constraints on each different attacker-victim pair. We propose an optimal method integrating within-step static constrained attack-resource allocation optimization and between-step dynamic programming to achieve the optimal adversarial attack in a multi-agent system. Our numerical results show that the proposed attacks can significantly reduce the rewards received by the attacked agents.
\end{abstract}
\section{Introduction}
\label{sec:intro}
Reinforcement learning (RL), including multi-agent reinforcement learning (MARL), is used as the underlying algorithm for agents to interact with and make decisions in a dynamic environment. In MARL, multiple players (agents) interact with the environment and, in turn,  the environment dynamics depend on the joint actions of all the agents. Researchers often assume that MARL is a Markov Decision Process (MDP) \cite{MDP} and the game model is a Markov game (MG) \cite{MG}.
RL has many applications in different scenarios,  for example, autonomous driving\cite{ad}, financial decisions\cite{fd}, recommendation systems\cite{rs}, and also in drones' and robots' algorithms \cite{drones}. Since RL models, including MARL, are increasingly used in critical tasks, understanding the security and adversarial robustness of RL and MARL is becoming more and more important.

There have been some recent works addressing adversarial attacks of single-agent RL \cite{davis-sa, enmy, enchat, unlearner, vul-mech}. There are in general 5 types of adversarial attacks: the observation poisoning attacks, the environment poisoning attacks, the action-only poisoning attack, reward-only poisoning attacks, and the mixed poisoning attacks \cite{att-costsig, davis-ma, ada-rpa,davis-sa, poli-induc, batch-rl}.
 For example, in\cite{enmy}, the authors investigated the optimality of evasion attacks (reduce the expected total reward gained by agents) under budget constraint from the perspective of policy perturbations during learning. In \cite{enchat}, the authors proposed two types of attacks, the strategically-timed attack and the enchanting attack, in which the attacker aims to lure the victim (recipient) agent into a designated target state.

For the MARL setting, most existing works investigated the model of one limitless attacker attacking multiple agents. In \cite{davis-ma}, with a generalized mediator's objective, the authors proposed a mediator model in which the mediator (attacker) attacks the agents in the game to mediate the agents to maximize the objective function in an MG. In \cite{commu}, the authors proposed adversarial attacks on the communication of multi-agent systems.
\cite{davis-ma} considers the existence of one limitless mediator (attacker) attacking recipient agents. However, in reality, attacks are more likely to be performed by distributed agents, and every attacker has a budget constraint on its resources used for attacks. 



In this paper, we propose a new formulation of adversarial attack in MARL: the agent-attack-agent model. In this model, distributed agents instead of central mediators perform adversarial attacks, each attack between each attacker and recipient pair has distinct costs, and each attacker is limited by its own attack budget. We propose an optimal method integrating within-step static constrained attack-resource allocation optimization and between-step dynamic programming to achieve the optimal adversarial attack in multiple agent systems. Our numerical results show that the proposed attacks can significantly reduce the rewards received by the attacked agents.

The rest of this paper is organized as follows. In Section \ref{sec:problem_form}, we introduce the agent-attack-agent model and formulate the problem of finding optimal attacks. In Section \ref{sec:dp_findoptatt}, we consider two budget constraint scenarios and propose two corresponding optimal algorithms for finding the optimal attack policy. In Section \ref{sec:num_results}, we numerically evaluate the performance of the proposed attack algorithms. 
\section{Problem Formulation}
\label{sec:problem_form}
In the considered MDP environment, all the agents are divided into two opposite groups, the attacker group $M$, and the recipient agent group $N$, with $|M| = m, \ |N| = n,$, and $m, n < \infty$.

The MDP environment for the recipient agents can be described by the 5-element tuple: $(\{\mathcal{S}_i\}_{i=1}^n, \{\mathcal{A}_i\}_{i=1}^n,$ $T, P, \{R_i\}_{i=1}^n)$, where $\mathcal{S}_i$ is the state space of the $i$-th recipient agent with $|\mathcal{S}_i|=S_i$, $\mathcal{A}_i$ is the action space for the $i$-th recipient agent with $|\mathcal{A}_i| = A_i$, $T$ is the number of time steps in MARL and $T$ is finite. Our time index $t$ starts from 0 to $T$. $P_{i,t}:\mathcal{S}_i \times \mathcal{A}_i \times \mathcal{S}_i \rightarrow [0,1]$ is the transitional probability of the $i$-th recipient agent at time index $t$, and ${R}_{i,t}$ represents the reward function of the $i$-th recipient agent at time index $t$. We refer to time step $t$ as the time interval starting from time index $t-1$ and ending at time index $t$, where $1\leq t \leq T$. We let $\textbf{a}_t := (a_1, a_2, \dots, a_n)$ denote the joint action of all the $n$ recipient agents at time step $t$, and let $\textbf{s}_t := (s_1, s_2, \dots, s_n)$ denote the joint state of all the $n$ recipient agents at a time index 
$t$. We define the optimal policy of the $i$-th recipient agent at time index  $t$ as $\pi_{i,t}^{\star}(s_{i,t}) = a_{i,t}^{\star}$. For simplicity, we assume that all the recipient agents share the same state spaces and the same action spaces.Therefore, the optimal policy $\pi_{i}^{\star}(s_{i,t})$ is the same for every agent $i$, denoted as $\pi_t^{\star}(s_{t})$. 

\textbf{State poisoning attack setting}: We assume both the attack and recipient agent groups can monitor the underlying MARL algorithms of the recipient agents and therefore both groups know the optimal policies $\pi^{\star}_{i,t}$ of every recipient agent $i$ at time step $t$. However, the recipient agents are unaware of the existence of attackers or their attacks. The $m$ attackers perform state poisoning attacks by disturbing recipient agents' observations of their true states. For a recipient agent $i$ at time index $t$,  we let $s_{a,t,i}$ denote the true state the agent $i$ is actually in and let $s_{d,t,i}$ denotes the delusional state that the agent $i$ thinks it is in. 



We consider $T$ time steps and the time index goes from 0 to $T$. The recipient agents are selfish in the game, aiming to maximize their own rewards obtained during the $T$ time steps (a time step $t$ is from the time index $t-1$ to $t$). The attackers instead aim to minimize the total expected rewards of all the recipient agents during the $T$ time steps.  

There are two phases of playing during one time step $t$. In the first phase, from time index $t-1$ to $t-0.5$, the attackers attack to make each recipient agent $i$ ($1\leq  i \leq n$) think it is in a delusional state ${s}_{d,t-0.5, i}$. In the 2nd phase, after the attack, from the time index $t-0.5$ to $t$, each recipient agent $i$ moves to $s_{a,t,i}$ according to its optimal policy $\pi_{i,t-0.5}^{\star}(s_{d,t-0.5,i})=\pi_{i,t-1}^{\star}(s_{d,t-0.5,i})$, in which $s_{d,t-0.5,i}$ is agent $i$'s delusional state at time index $t-0.5$, and obtains its corresponding reward $R_{i,t}(s_{a,t-1,i}, s_{a,t,i})$.

We are interested in finding the optimal attack strategy of every attack agent for each time step $t$ ($1\leq t \leq T$).

\section{Finding the Optimal Attack with Agent-to-Agent Attack Cost Constraints}
\label{sec:dp_findoptatt}
We look at two cases where the attackers are constrained by attack budgets and there are different costs associated with each attacker-recipient pair. 

\subsection{All-time cost constrained case}
\label{sssec:al-time-cc}
At time index $0$, each attacker is assigned an attack budget that is available for spending during the whole game. We call this scenario an ``all-time constrained cost case". At $t=0$, each attacker $j$ is assigned its own budget $b_j$ and this budget $b_j$ can be spent freely before time index $T$. Namely, if there is still a budget left for agent $i$, the leftover budget can be used in later time steps (as long as before time index $T$).

We design a dynamic programming between steps to compute such an optimal attack strategy for attackers. For each time index $t$ ($0\leq t \leq T$), we define a state in the dynamic programming (called Dynamic Programming State (DPS)) as $\sigma_t = (\textbf{s}_{a,t}, \textbf{s}_{d,t}, \textbf{b}_t)$, where $\textbf{s}_{a,t}$ are the true states of recipient agents at the time index $t$, $\textbf{s}_{d,t}$ are the delusional states the recipient agents think they are in at the time index $t$, and $\textbf{b}_t$ are the amount of unspent budgets of attackers at the time index $t$. From the two-phase model, for any integer $t$ between $1$ and $T$,  we have $\textbf{s}_{a,t-0.5}=\textbf{s}_{a,t-1}$ (the attacks do not change the actual states), and  $\textbf{s}_{d,t}=\textbf{s}_{d,t-0.5}$ (actually we can take $\textbf{s}_{d,t}$ at integer indices $t$ to be any value, because the recipient agents take actions only based on delusional states at 0.5-time-indices between two integers). We denote the set of DPS states as $\boldsymbol\Sigma$.
\begin{figure}[htb]
  \centering
  \centerline{\includegraphics[width=8cm]{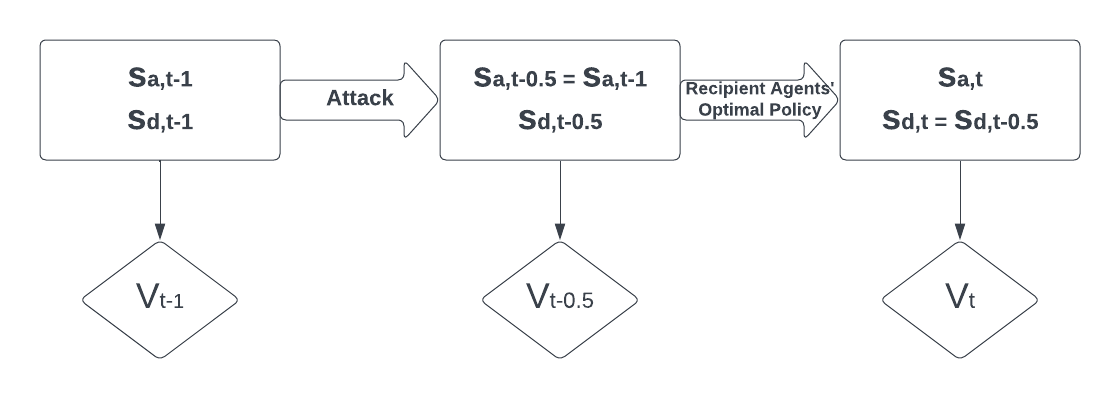}}
  \caption{The transitions between states}

\end{figure}

We define the set of joint attack actions (performed by all the $m$ attackers) as $\mathcal{A}^{\dagger}$, and denote a joint attack happening at integer time index $t$ as $a^{\dagger}_t = (\textbf{s}_{g,t}, \boldsymbol{\beta}_t)$, where $\textbf{s}_{g,t} \in \mathcal{S}^n$. Here $\mathbf{s}_{g,t,i}$ represents the target delusional states the attackers are trying to fool the agent $i$ into, and $\boldsymbol{\beta}_t$ is an $n \times m$ matrix with $0$ or $1$ as its elements. If $(\boldsymbol{\beta}_t)_{i,j}=1$, it denotes attacker $j$ participates in attacking recipient agent $i$ and spends corresponding attack resources $D(i, j)$; otherwise it does not attack recipient agent $i$. Our goal is to find the optimal attack $a^{\dagger,\star}_t$ at each time index $t$.

Thus, for every DPS state $\sigma_t$ at time index $t$, we introduce the dynamic programming state values (DPSV) at time index $t$ under attack $a^{\dagger}_t$ as $V_t(\sigma_t, a^{\dagger}_t)$.  This value records the expected total rewards received by all recipient agents from time index $t$ to time index $T$ if the joint attack at time $t$ is $a^{\dagger}_t$. We call $V^{\star}_t(\sigma_t)$ the Minimal Possible Value (MPV) of the state $\sigma_t$ at time index $t$. In dynamic programming, we update it as 
$V^{\star}_t(\sigma_t)=\min_{a^{\dagger}_t \in \mathcal{A}^{\dagger}}V_t(\sigma_t, a^{\dagger}_t)$. The corresponding optimal attack at time index $t$ is given by $\arg \min_{a_t^{\dagger}\in \mathcal{A}^{\dagger}}V_t(\sigma_t, a_t^{\dagger})$, denoted as $a^{\dagger,\star}_t$. Note that we update the leftover budgets in transitions between states. The algorithm is described as follows.



\begin{figure}[htb]

\begin{minipage}[b]{1.0\linewidth}
  \centering
  \centerline{\includegraphics[width=9.4cm]{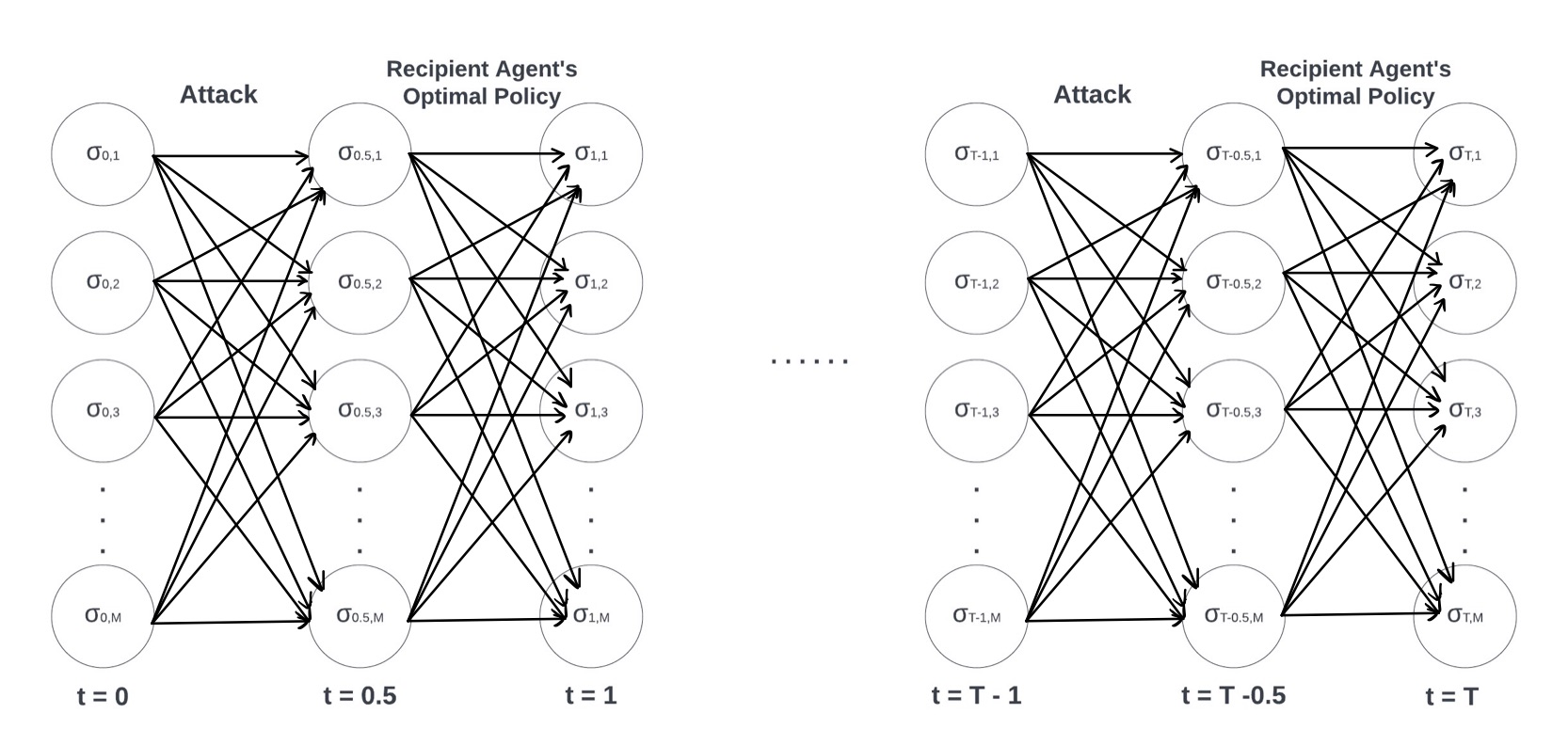}}
  \caption{An illustration of dynamic programming for computing optimal attacks, where $M$ is the number of DPS's.}
\end{minipage}

\end{figure}

\SetKwComment{Comment}{/* }{ */}

\begin{algorithm}[hbt!]
\caption{All-time constrained optimal attack}\label{alg:two}
\KwData{{DPS space $\boldsymbol\Sigma$, attack space $\mathcal{A}^{\dagger}$, recipients' optimal policy $\pi^{\star}$, recipients' reward function $R$, transition probability $P$ under recipient action, transition probability $P'$ under attack}}
\KwResult{\small{Optimal attacks $a^{\dagger, \star}_t$, MPV $V_t^{\star}(\sigma)$, $0\leq t\leq T$}}\
Initialize $V_T^{\star}(\sigma_{T}) \gets 0, \forall \sigma_{T} \in \Sigma$;\\
\For{$t = T, T-2, \dots, 1$}{
\For{every $\sigma_t$ in $\boldsymbol\Sigma$}{
$V_{t-0.5}^{\star}(\sigma_{t-0.5}) = \sum_{\sigma_t} P(\sigma_{t}|\sigma_{t-0.5}, \textbf{a}^{\star}_t) ( V_t^{\star}(\sigma_t)  + R(\sigma_{t-0.5},\sigma_t))$\;
\For {available $a^{\dagger}_{t-1}$ in $\mathcal{A}^{\dagger}$}{
$V_{t-1}(\sigma_{t-1}, a^{\dagger}_{t-1}) = \sum_{\sigma_{t-0.5}} P'(\sigma_{t-0.5}|\sigma_{t-1}, a^{\dagger}_{t-1}) V_{t-0.5}^{\star}(\sigma_{t-0.5})$
}
$V_{t-1}^{\star}(\sigma_{t-1}) = \min_{a^{\dagger}_{t-1}} V_{t-1}(\sigma_{t-1}, a^{\dagger}_{t-1})$\;
$a^{\dagger, \star}_{t-1} = \arg \min_{a^{\dagger}_{t-1}}  V_{t-1}(\sigma_{t-1}, a^{\dagger}_{t-1})$
}
}
\end{algorithm}



\subsection{Instant cost constrained case}
\label{sssec:al-time-cc}

In practice, sometimes the attackers have attack budgets that must be spent by the end of each time step. For example, the resources used by attackers are provided by constantly-energy-harvesting systems over time steps, and the budget for each time step is constrained by the battery volume.  We call this scenario an ``instant cost constrained case''. Within each time step $t$ (namely between time index $t-1$ and $t$),  each attacker $j$ has its own budget $b_j$ and this budget $b_j$ can only be spent during that single time step: the leftover resources cannot be carried over to the next time step $t+1$ or there is no need to carry over the leftover resources to the next time step because of budget refill. Once we get to time step $t+1$, every attacker's (say, attacker $j$) budget will be refilled (say, to $b_j$). We would like to find out how to optimally allocate each attacker's resources to attacking each recipient for each step (from time step 1 to step $T$), while satisfying the instant cost constraint and minimizing the recipient agents' total rewards. 

We design an integration of between-step dynamic programming and within-step static constrained optimizations to compute the optimal attack strategy. So during each time step $t$, for each possible actual state vector $\textbf{s}_{a,t-1}$, we propose to use a static standalone optimization program to determine the optimal allocation of attackers' budgets on attacking recipients.  Between different time steps, we use dynamic programming to account for the state transitions and expected rewards.

We work backward from time index $t=T$ and initialize $V_{T}^{\star}(\sigma_T)=0$ for each DPS $\sigma_{T}$ (a DPS state includes all recipient agents' actual states and delusional states), and the subscript represents time index. Suppose that we have already computed $V_{t+0.5}^{\star}(\sigma_{t+0.5})$ for every DPS state $\sigma_{t+0.5}$.  We then proceed to compute the optimal attack policy during time step $t$ (essentially from $t$ to $t+0.5$) and also $V_{t}^{\star}(\sigma_t)$ for every DPS state $\sigma_t$. During time step $t$, we let $x_{ij} \in \mathbb{R}$ be the amount of resources attacker $j$ spends on attacking recipient agent $i$. The constraints on $x_{ij}$ are such that the total spending of each attacker (each attacker can attack multiple recipient agents) cannot exceed its respective budget. To represent formulas concisely, we stack the $x_{ij}$'s to form a $mn$-dimensional vector $\mathbf{x}$ called the attack allocation vector. Under the attack allocation vector $\mathbf{x}$, we denote the probability that the recipient agents' state will transition to $\sigma_{t+0.5}$ as $P(\mathbf{x},\mathbf{s}_{a,t}, \sigma_{t+0.5})$, where $\mathbf{s}_{a,t}$ is the true states of all the recipient agents at time index $t$. 
This probability must be between 0 and 1. Based on the principles of dynamic programming, we want to optimize $x_{ij}$'s to minimize the total expected rewards received by the agents from time step $t$ to $T$. Thus, under a specific true state vector $\mathbf{s}_{a,t}$, the objective function for attackers to minimize is the expected total reward of all the recipient agents from step $t$ onward to step $T$. 
\subsubsection{Within-step static constrained optimization problem}
\label{sec:staticoptimization}
Noticing that the DPS has $S^n$ (assuming each recipient agent has $S$ states) possible values at time index $t+0.5$ conditioned on the true states are $\mathbf{s}_{a,t}$, the optimal attack under the ``instant cost constrained'' case at a single time step $t$ can be formulated as the following within-step static constrained optimization problem:
\begin{align}
\label{static}
\min \sum_{k=1}^{S^{n}} & P(\mathbf{x},\mathbf{s}_{a,t}, \sigma_{t+0.5}^k) V_{t+0.5}^{\star}(\sigma_{t+0.5}^k)\\
\text{subject to} &\sum_{i=1}^{n} x_{ij} \leq b_j, \  j = 1,\dots,m \nonumber\\
& P(\mathbf{x},\mathbf{s}_{a,t}, \sigma_{t+0.5}^k) \leq 1, \ \forall k  \nonumber \\
& -P(\mathbf{x},\mathbf{s}_{a,t}, \sigma_{t+0.5}^k) \leq 0, \  \forall k \nonumber\\
& x_{ij} \geq 0, \ i=1,\dots,n, \  j = 1,\dots,m, \nonumber
\end{align}
where $\sigma_{t+0.5}^k$ is the $k$-th DPS at time index $t+0.5$.

Depending on the physical nature of the attacks, we can model the probability $P(\mathbf{x},\mathbf{s}_{a,t}, \sigma_{t+0.5}^k)$ as different functions of $\mathbf{x}$. In one particular ``proportional effort success'' model considered in the paper, for each recipient agent $i$, the probability that its delusional state will be different from its true state is $\max\{\sum_{j=1}^{m} [x_{ij}/C_t(i,j)],1\}$, where $C_t(i,j)$ are constants representing how hard it is for attacker $j$ to attack recipient agent $i$.  Namely, that probability grows with $\sum_{j=1}^{m} [x_{ij}/C_t(i,j)]$ until $\sum_{j=1}^{m} x_{ij}/C_i(i,j)$ hits $1$. We also assume that conditioned on the delusional state is different from the actual state, the delusional state is any feasible state (namely a state distinct from the actual state) with equal probability.  In our numerical results, we take $C_t(i,j)=[d^t_{ij}]^3+\epsilon$, where $\epsilon$ is positive constant and $d^t_{ij}$ is the distance between attacker $j$ and recipient agent $i$ at time index $t$. In numerical results, we also assume that $P(\mathbf{x},\mathbf{s}_{a,t}, \sigma_{t+0.5}^k)$ is such that different recipient agents take delusional state values independently from each other.  

\subsubsection{Between-step dynamic programming}
After solving \eqref{static}, we take the optimal value of its objective function as $V_t^{\star}(\sigma)$, using which we continue to calculate $V_{t-0.5}^{\star}(\cdot)$ as follows. For each DPS $\sigma_{t-0.5}$, we update $V_{t-0.5}^{\star}(\sigma_{t-0.5})$ as 
$$\small \sum_{\sigma_t} P(\sigma_{t} | \sigma_{t-0.5}, \textbf{a}^{\star}_{t-0.5}) ( V_t^{\star}(\sigma_t)  + R(\sigma_{t-0.5},\sigma_t)).$$
After updating $V_{t-0.5}^{\star}(\cdot)$, again we will use another static optimization formulation in Section \ref{sec:staticoptimization} to calculate $V_{t-1}^{\star}(\cdot)$. In this way, we perform this static optimization-dynamic programming cycle recursively until we calculate all the $V_{t}^{\star}(\cdot)$ backward from $t=T$ until $t=0$. 


\section{Numerical Results}
\label{sec:num_results}
We perform numerical simulations with $T=5$. There are 2 recipient agents and 2 attackers playing in this MG. Both recipients share the same state space $S$, action space $A$, the probability transition $P$, and reward function $R$. The three states correspond to three locations on a circle. The action space $A$ is composed of three actions: go left, go right, and stay. For actions \textbf{left} and \textbf{right}, the recipient agent has a 0.8 probability of moving in the intended direction and a 0.2 probability of moving in the opposite direction. For \textbf{stay}, the recipient agent has a 0.8 probability of staying at the current state and a 0.1 possibility of moving to the right or left. 

\subsection{All-time cost constrained case}
We assign $b_1 = 10$ and $b_2 = 10$ at time index $t=0$, meaning the total attack spending of each attacker cannot exceed 10. The attacker-victim pair attack costs are $D(1,1) = 3$, $D(2,1)=2$, $D(1,2)=1$ and $D(2,2)=1$. In Figure 3, we compare the expected global rewards (benefits) of recipient agents as the time index goes from $t=0$ to $t=5$,  between our optimal attack and a random attack where in each time step the two attackers randomly pick an attack action. 

\subsection{Instant cost constrained case}
In this case, within every time step $t$, budget $b_1$ is refilled to 0.8, and $b_2$ is refilled to 0.5. Attackers' locations are fixed:  attacker 1 at state (location) 0 and attacker 2 at state 2. We use the earlier mentioned ``proportional effort success'' model to relate attack spending with the probability of getting into a  delusional state, with $\epsilon=0.0001$.  
In Figure 4, we compare the optimal attack with the following random attack: attacker 1 allocates its entire budget to a random recipient agent, while attacker 2 allocates its entire budget toward attacking the other agent.  


\begin{figure}[htb]
\begin{minipage}[b]{0.48\linewidth}
  \centering
  \centerline{\includegraphics[width=4.4cm, height=4.5cm]{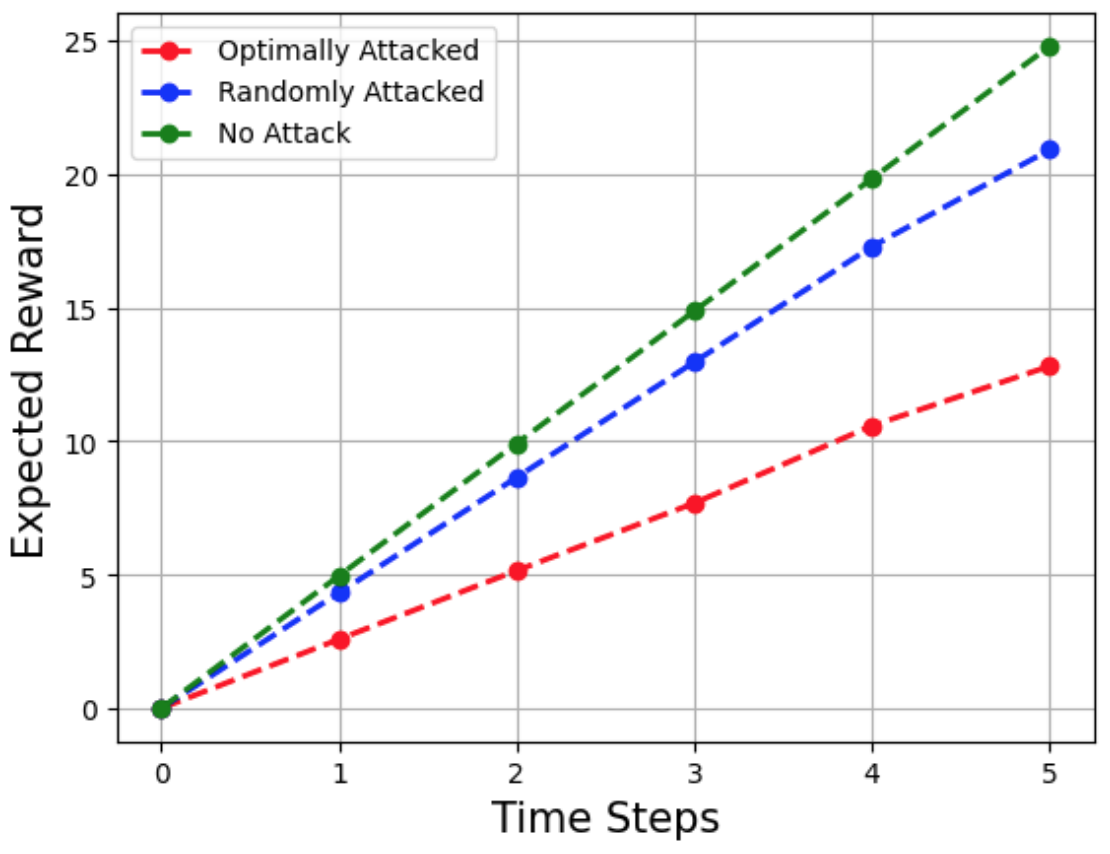}}
  \centering{\textbf{Fig. 3.} All-time cost-constrained case}
\end{minipage}
\hfill
\begin{minipage}[b]{0.48\linewidth}
  \centering
  \centerline{\includegraphics[width=4.4cm, height=4.5cm]{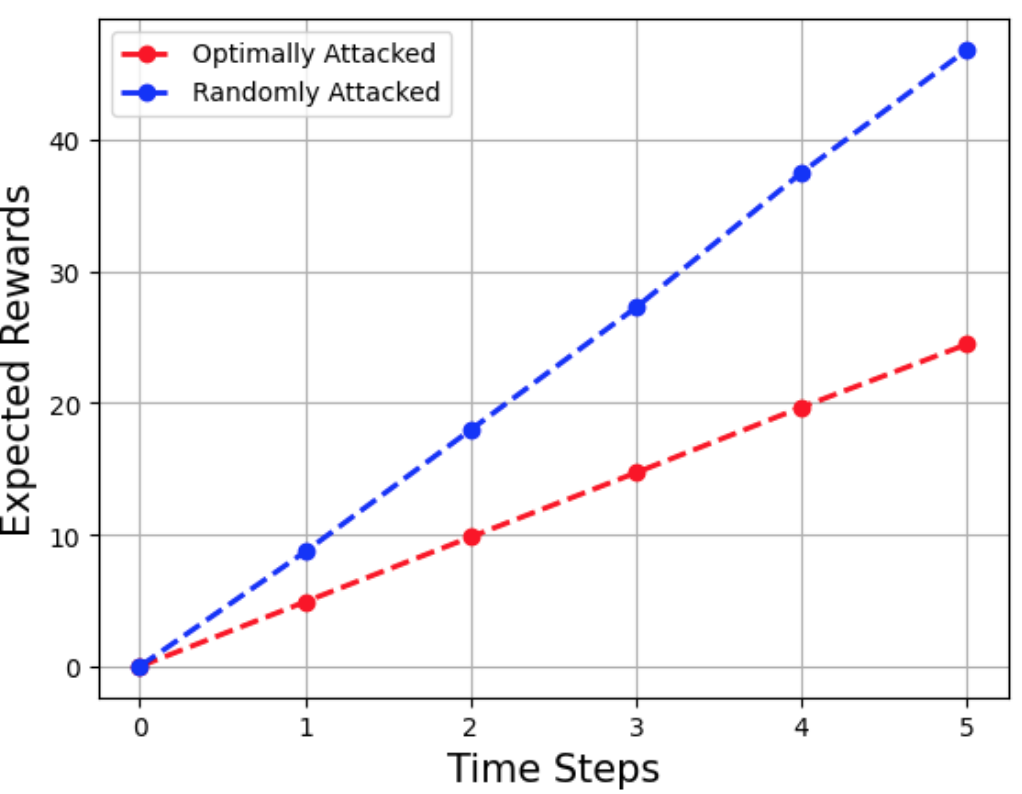}}
  \centering{\textbf{Fig. 4.} Instant cost-constrained case}
\end{minipage}
\label{fig:res}
\end{figure}
In both figures, the x-axis is the current time index, and the y-axis is the total expected rewards the recipient agents obtained starting from $t=0$ until that current time index. In the all-time cost constrained case, with optimal attacks, the reward gain is roughly 61$\%$ of that achieved through the random attack and 51$\%$ of that achieved without attack. In the instant cost-constrained case, the optimal attack achieves about 52$\%$ of the reward obtained under random attacks and 28$\%$ of the reward without attacks.

\vfill\pagebreak

\label{sec:refs}

\bibliographystyle{IEEEbib}
\bibliography{refs}

\end{document}